\documentclass[conference]{IEEEtran}
\IEEEoverridecommandlockouts
\usepackage{cite}
\usepackage{amsmath,amssymb,amsfonts}
\usepackage{algorithmic}
\usepackage{graphicx}
\usepackage{textcomp}
\usepackage{xcolor}
\def\BibTeX{{\rm B\kern-.05em{\sc i\kern-.025em b}\kern-.08em
    T\kern-.1667em\lower.7ex\hbox{E}\kern-.125emX}}

\usepackage{multirow}
\usepackage{float}
\usepackage{balance}
    
\begin{document}

\title{User Identification across Social Networking Sites using User Profiles and Posting Patterns}

\author{\IEEEauthorblockN{Prashant Solanki\IEEEauthorrefmark{1}, Kwan Hui Lim\IEEEauthorrefmark{2} and Aaron Harwood\IEEEauthorrefmark{1}}
\IEEEauthorblockA{\IEEEauthorrefmark{1}School of Computing and Information Systems, The University of Melbourne\\}
\IEEEauthorblockA{\IEEEauthorrefmark{2}Information Systems Technology and Design Pillar, Singapore University of Technology and Design\\}
\IEEEauthorblockA{Email: solankip@student.unimelb.edu.au, kwanhui\_lim@sutd.edu.sg, aharwood@unimelb.edu.au}
}


\maketitle

\begin{abstract}
With the prevalence of online social networking sites (OSNs) and mobile devices, people are increasingly reliant on a variety of OSNs for keeping in touch with family and friends, and using it as a source of information. For example, a user might utilise multiple OSNs for different purposes, such as using Flickr to share holiday pictures with family and friends, and Twitter to post short messages about their thoughts. Identifying the same user across multiple OSNs is an important task as this allows us to understand the usage patterns of users among different OSNs, make recommendations when a user registers for a new OSN, and various other useful applications. To address this problem, we proposed an algorithm based on the multilayer perceptron using various types of features, namely: (i) user profile, such as name, location, description; (ii) temporal distribution of user generated content; and (iii) embedding based on user name, real name and description. Using a Twitter and Flickr dataset of users and their posting activities, we perform an empirical study on how these features affect the performance of user identification across the two OSNs and discuss our main findings based on the different features.
\end{abstract}


\section{Introduction}

Online social networking sites (OSNs) are increasingly prevalent in our everyday life and this trend has been more pronounced in recent times due to the social distancing measures brought about by the COVID-19 pandemic~\cite{cellini2020changes,Kwan-ASONAM20,goel2020social,chan2020social}. As such, people are increasingly reliant on a variety of OSNs for keeping in touch with family and friends, and using it as a source of information. For instance, a user might use Flickr to share photos about their daily lives with family and friends, while using Twitter to post short messages that reflect their opinions and thoughts about various issues.

With the use of multiple OSNs, an important problem is the identification of the same user across these OSNs as this allows us to understand the usage patterns of users among different OSNs, make recommendations when a user registers for a new OSN, and various other useful applications~\cite{jain2013seek,shu2017user}. In this work, we study the important problem of user identification across multiple, different OSNs. We make the following main contributions:
\begin{itemize}
\item We propose three models, based on a multi-layer perceptron architecture, for the task of user identification across OSNs, namely: (i) Profile Similarity model that uses various user profile features coupled with similarity measures (Section~\ref{profile_similarity_model}); (ii) Temporal Model based on temporal patterns in the posting of UGCs (Section~\ref{sectTemporalModel}); and (iii) Profile Embedding model that utilizes word/character embedding techniques on textual features in the user profile (Section~\ref{sectProfileEmbedModel}).
\item We collected and analyzed 75,398 user pairs between Twitter and LinkedIn, Instagram, Flickr, Reddit, Tumblr and Pinterest, before selecting Twitter-Flickr as our cross OSN dataset. Thereafter, we build upon an existing dataset of Twitter-Flickr user pairs~\cite{Cosnet2015} and further augment this dataset by collecting additional information relating to the user profiles and posting patterns. This augmented dataset comprises 125,586 tweets and 71,784 Flickr posts from 955 Twitter-Flickr user pairs (Section~\ref{data_description}).
\item We empirically evaluate the effectiveness of each model, including a comprehensive analysis of: (i) five user profile features and nine similarity measures in the Profile Similarity model; (ii) temporal posting of user content based on daily and weekly patterns in the Temporal model; (iii) Using word and character embedding techniques alongside similarity measures for the Profile Embedding model. Experiment results show that the Profile Similarity model performs the best, followed by the Profile Embedding model and Temporal model, based on Precision, Recall and F1-score (Section~\ref{sectExperimentResults}).
\end{itemize}

\section{Background and Related Work}

In this section, we discuss related work on user identification using a variety of features.

\subsection{User Identification based on User Profile Features}

Data privacy is a concern among OSN users as users may provide their personal information to multiple OSNs but do not intend for this data to be integrated \cite{Peled2016}. Social networks have data about each user's profile containing information such as - age, location, interests, pictures, work, and education; his or her social links (friends); and a variety of additional services, such as email IDs and instant messaging IDs. \cite{Peled2016} worked on user de-anonymizing that involves finding a user's hidden account (under a pseudonym) on an OSN, given his name and profile details on another OSN.

\cite{Ma2016} aims to provide holistic approach of collecting data across the ever expanding pool of OSNs. Faced with the issue of scalability and poor results of blind string comparisons, the authors have developed a profile discovery schema named \textbf{Splicer}. The tool is primarily based upon three types of similarity functions, \textit{exact}, \textit{partial} and \textit{statistical}. The results obtained from these three functions are deployed to come up with the pair-wise profile joining. \cite{Ma2016} created a scalable solution that discover heterogeneous information by reducing computational time by 87\%.  

\cite{Goga2015} proposed a solution which is centred around properties of profile attributes like Availability, Consistency, non-Impersonation, and Discrimination. These properties are essential to determine the reliability of the matching schema. The solution talks about the limitations faced in comparing the real world profiles and thus showcase the methodologies which can be employed in realistic situations.
The authors reported a Recall result of 90{\%} when a small dataset was employed to train the algorithm but it significantly dropped to 19{\%} when a large dataset was brought into view.

Online Social Networks have become overly populated in the last decade and the tally is expected to reach 2.95 billion until 2020. Also, it is a common practice to use different social networking sites for various purposes. \cite{Li2017} highlights the importance of integrating the data obtained in these sites to build a complete picture of an individual. 
The authors analysed the redundancy in the display name of the user, exploited this information via various features and proposed a logistic regression classifier based various of these features.

Rich user profile data across various social networking sites is the major concept encircled by different scholars. This is adversely affected by either the reluctance of user in providing complete information or various privacy settings. \cite{Li2018a} aimed to overcome this issue by using User Generated Content (UGCs) to accomplish the goal of User identification.  
The authors proposed a triple layered classifier by cascading the results of machine learning algorithms based on temporal, spatial and content features.

\subsection{User Identification based on Temporal, Spatial and Network Features}

Many OSN users utilize services like If This Then That (IFTTT) that enables them to connect and simultaneously post to multiple OSNs. Researchers like \cite{Roedler2017} studied aspects like the posting time difference between the post of Instagram and the corresponding tweet. \cite{Roedler2017} observes that 75\% of the Instagram links were tweeted within 3 seconds of posting on Instagram and hypothesized that such small timeframes can only be achieved when using an automated approach like IFTTT.

Location data collected from user devices are widely used by application to a make personalized recommendations and advertisements. This data is also highly valuable for identifying an individual on multiple social networks, as the location information along with timestamps creates a unique signature of the user. \cite{Roedler2016} studied usage and geo-tagging behaviour in relation to time and places, and observed that time and location of usage of an OSN rarely varies for an individual.

Using only location tags to identify the user is dependent on user attaching the location of posting in the post/tweet, which might not always be the case. According to \cite{Roedler2016}, only 31\% of the users from the total collection of tweets used location tags. Out of these 31\% users, only 17\% shared their locations four or more times a day, which is important for sequence detection. 

The basis of the work carried out in \cite{GurcanAkcora} is based on user similarities based on their connections and profile attributes. Their study includes datasets from Facebook and DBLP which is enriched with profile data from ACM digital library. They utilized similarity measures based on network and profile similarities. These similarity measures are derived from real-life usage patterns like users become friends with people who are either close to them and/or have similar profile attributes like gender, education and religion.

\section{Proposed Models}
\label{sectProposedModels}

In this section, we describe the various models we propose for identifying users across multiple social networks.

\subsection{Profile Similarity (PS) model - User Identification based on Profile Similarity}
\label{profile_similarity_model}

The PS model aims to identify users using their public profile. For an individual's social account on a OSN (e.g., Twitter), we examine various user profile information (e.g., name of the person, screen name, location, etc) and try to match the account of this individual at another OSN (e.g., Flickr). For this model, we first extract features based on the various text field present in the user profiles, as shown in Table~\ref{tab:model:pfs:featuresSimilarity}. 

Thereafter, these features were then calculated with 9 different similarity measures, as shown in Table~\ref{tab:model:pfs:featuresSimilarity}. Given any two user profiles $U^{i=[1..5]}_a$ and $U^{i=[1..5]}_b$, where $i$ is the extracted feature ranging from 1 to 5 and $a,b$ are the two OSN profiles for which we computed similarities $S^j = Sim^j(U^{i=[1..5]}_a,U^{i=[1..5]}_b)$ where $j$ is the normalized similarities given in Table~\ref{tab:model:pfs:featuresSimilarity}. In the following sections, we further describe each of these similarity measures.

\begin{table}
	\renewcommand*{\arraystretch}{1.1}
\centering
\caption{Features and similarity measures used in the Profile Similarity model}
\label{tab:model:pfs:featuresSimilarity} 
\begin{tabular}{ cc } 
\hline
Feature & Similarity Measure\\
\hline
& Levenshtein\\
& Damerau-Levenshtein\\
User Name Score & Editex\\
Real Name Score & Jaro–Winkler\\
Post Ratio & Jaccard\\
Description Score & bzip2\\
Location Score & Longest Common Subsequence\\
& Smith-Waterman\\
& Cosine\\
\hline
\end{tabular}
\end{table}

\subsubsection{Levenshtein distance}
The Levenshtein distance is a minimum number of single-character edits like insertions, deletions or substitutions required to change one word into the other word \cite{Levenshtein1966}.
For two strings $a$ and $b$, this is defined as:
\begin{figure}[h]
    \centering
{${lev} _{a,b}(i,j)={\begin{cases}
\max(i,j)&{\text{ if }}\min(i,j)=0,\\
\min {\begin{cases}
\operatorname {lev} _{a,b}(i-1,j)+1\\
\operatorname {lev} _{a,b}(i,j-1)+1\\
\operatorname {lev} _{a,b}(i-1,j-1)+1_{(a_{i}\neq b_{j})}
\end{cases}}
&{\text{ otherwise.}}
\end{cases}}$}
    \label{fig:levenshtein_distance}
\end{figure}

where $1_{(a_{i}\neq b_{j})}$ is equal to 0 when ${a_{i}=b_{j}}$ and equal to 1 when ${a_{i} \neq b_{j}}$, and ${\operatorname {lev} _{a,b}(i,j)}$ shows the distance between the first $i_{th}$ characters of string $a$ and the first $j_{th}$ characters of string $b$. 

\subsubsection{Damerau-Levenshtein distance}

The Damerau-Levenshtein distance is a revision of Levenshtein distance with a comprehensive set of operations comprising of insertions, deletions or substitutions of a single character, or transposition of two adjacent characters\cite{Leven}. The distance is calculated by the minimum number of edits required to change one word into another word, defined recursively as: 

\begin{figure}[H]
    \centering
 ${d_{a,b}(i,j)=}$ 
 $\min {\begin{cases}0&{\text{if }}i=j=0\\d_{a,b}(i-1,j)+1&{\text{if }}i>0\\d_{a,b}(i,j-1)+1&{\text{if }}j>0\\d_{a,b}(i-1,j-1)+1_{(a_{i}\neq b_{j})}&{\text{if }}i,j>0\\d_{a,b}(i-2,j-2)+1&{\text{if }}i,j>1\\&{\text{ and }}a[i]=b[j-1]\\&{\text{ and }}a[i-1]=b[j]\\\end{cases}}$
    \label{fig:damerau_levenshtein}
\end{figure}

where $i$ and $j$ denotes the length of the string $a$ and $b$ respectively.
Each of the cases represent a different operation: (i) $d_{a,b}(i-1,j)+1$: deletion; (ii) $d_{a,b}(i,j-1)+1$ : insertion; (iii) $d_{a,b}(i-1,j-1)+1_{(a_{i}\neq b_{j})}$: match or mismatch; (iv) $d_{a,b}(i-2,j-2)+1$: transposition between 2 characters.

\subsubsection{Editex Similarity}
Editex is a phonetic matching algorithm which is an improved adaptation of the Soundex algorithm. Phonetic matching is a highly exercised method utilised to identify similar sounding strings like usernames or name of companies \cite{Zobel}. Editex merges the properties of Soundex, Phonix and edits distance. Editex utilises the letter grouping from Soundex and combines it with edit distance to get a better match than any one of the methods. 

\begin{figure}[H]
    \centering
$
edit(i,j) = {
    \begin{cases}
        0&{\text{if }} i=0,j=0,\\
        edit(i-1,0) + d (s_{i-1},s_i)&{\text{if }} j=0,\\
        edit(0,j-1) + d(t_{j-1},t_j)&{\text{if }} i=0,\\
        \min{
            \begin{cases}
                edit(i-1,j) + d(s_{i-1},s_i)\\
                edit(i,j-1) + d(t_{j-1},t_j)\\
                edit(i-1,j-1) + r(s_i,t_j)\\
            \end{cases}
            }
            & {\text{otherwise}}
    \end{cases}
    }
$
    \label{fig:editex_similarity}
\end{figure}

\begin{table}[H]
	\renewcommand*{\arraystretch}{1.1}
    \centering
    \caption{Editex Letter Groups}
    \label{tab:editex_letter_groups}
    \begin{tabular}{ccccccccccc}
    \hline
        Code&0&1&2&3&4&5&6&7&8&9\\
    \hline
        Letters&aeiouy&bp&ckq&dt&lr&mn&gj&fpv&sxz&csz\\
    \hline
    \end{tabular}
\end{table}

Editex is based on the edit distance recurrence given in the above equation 
with a redefined function $r(i,j)$ which returns 0 if $i=j$, 1 if $i,j \in G$ , where G represents the group of letters in Table \ref{tab:editex_letter_groups} and 2 otherwise \cite{Zobel}. Apart from that Editex defines an additional function $d(i,j)$ which is similar to $r(a,b)$ except for one case, i.e. if $i= h$ or $i= w$ (letters that are often silent) and $i \neq j$ then $d(i,j)$ returns 1.

\subsubsection{Cosine Similarity}
The cosine similarity between any two vectors is calculated by finding the cosine of the angle between the two vectors \cite{Cosine}. 
Mathematically, cosine similarity can be calculated by solving the equation of $\cos{\theta}$.
\begin{figure}[H]
    \centering
$\vec{a} \cdot \vec{b} = \|\vec{a}\|\|\vec{b}\|\cos{\theta}$ \\
$\cos{\theta} = \frac{\vec{a} \cdot \vec{b}}{\|\vec{a}\|\|\vec{b}\|} $
    \label{fig:cosine_similarity}
\end{figure}

where, $\vec{a}$ and $\vec{b}$ represent the two vectors.

\subsubsection{Jaccard Similarity Coefficient}
\label{section:jaccard_similarity}
Jaccard similarity is a measure of how similar two sets are, based on the intersection of the sets over the union of the sets\cite{Jaccard}. 
In our case, the Jaccard similarity is calculated by making a set of 2-grams of the text from the data of user obtained from the two OSNs and finding the intersection and union of the two sets.

\begin{figure}[H]
    \centering
    $ J(A,B) = {{|A \cap B|}\over{|A \cup B|}}$
    \label{fig:jaccard_similarity}
\end{figure}

Jaccard Similarity lies between 0 and 1, where 0 being completely different and 1 being identical.

\subsubsection{Normalised compression distance (NCD) based on Bzip2}

NCD is a parameter-free metric of measuring the similarity between any two objects, images, texts, documents, music files, to name a few \cite{Vitanyi2011}. NCD does not require any features or background knowledge about the data to find the distance between two objects. NCD can be calculated by utilizing any of the normal compression techniques, in our study we used Bzip2 compression to calculate NCD of the textual features of the user. 

The following equation represents the mathematical notation of NCD.
\begin{figure}[H]
    \centering
$ NCD_{C}(a,b)={\frac  {C(ab)-\min\{C(a),C(b)\}}{\max\{C(a),C(b)\}}} $
    \label{fig:ncd_bzip2}
\end{figure}

where $C$ is a normal compression algorithm, is our case Bzip2. $a$ and $b$ are the two objects which are subjected to compression. NCD gives us the distance between the objects $a$ and $b$.
 
\subsubsection{Longest Common Subsequence (LCS)}
A sequence that is extracted from another sequence by the elimination of a few of the elements, not changing the order of the rest of the elements is called a subsequence. Longest common subsequence (LCS) of two strings or sequences is a subsequence of maximum length with can be found in both the sequences. LCS is defined as:

\begin{figure}[H]
    \centering
${
{{LCS}} (X_{i},Y_{j})=
}$
${
{\begin{cases}
\emptyset &{\mbox{if }}i=0{\mbox{ or }}j=0\\
{{LCS}}(X_{i-1},Y_{j-1}){\hat {}}x_{i}&{\mbox{if }}i,j>0\\&{\mbox{ and }}x_{i}=y_{j}\\
\operatorname {\max } \{{{LCS}}(X_{i},Y_{j-1}),{{LCS}}(X_{i-1},Y_{j})\}&
{\mbox{if }}i,j>0\\&{\mbox{ and }}x_{i}\neq y_{j}.\end{cases}}
}$
    \label{fig:longest_common_subsequence}
\end{figure}

where two sequences are represented as $X=(x_{1}x_{2}\cdots x_{m})$ and $Y=(y_{1}y_{2}\cdots y_{n})$. ${LCS}(X_{i},Y_{j})$ defines the LCS of $X_{i}$ and $Y_j$.
LCS gives use the longest subsequence between any two sequences and therefore, can be used a metric of the similarity between them. The longer length of LCS denotes more similarity between the two sequences. 

\subsubsection{Smith Waterman algorithm}
Smith-Waterman algorithm compares any two sequences by scoring the matches, insertions, and deletion required to make the strings identical~\cite{smith1981identification}. The Smith-Waterman algorithm is a four step process:

\begin{enumerate}
    \item \textbf{Defining the substitution matrix and the gap penalty function}.  
    Substitution matrix is created to assign the sequences a score for matching/mismatching.The gap penalty function is used to score the cost of gaps.    
    \item \textbf{Initialising the scoring matrix}. 
    Scoring matrix is 1 + the length of each sequences, with first row and column initialised with 0. 
    \item \textbf{Scoring}. 
    Scoring each character of the sequence starts from left to right, top to bottom and scores are recorded based on the operation like for substitution scores are diagonal and for adding gaps the scores are vertical or horizontal. If an element has all negative scores, then the element's score it set to 0, otherwise the score is set to the highest value and the sources of that score are recorded for tracing back.    
    \item \textbf{Tracing back}.
    Tracing back starts is initialised at the highest scored element from where the source of that element is traced back recursively until we hit a 0.
\end{enumerate}

The sequence with the highest similarity score is generated by the process of tracing back.

\subsection{Temporal Model - User Identification based on User-Generated Content (UGC)}
\label{sectTemporalModel}
The motivation behind this model was to determine the posting patterns of the users. For instance, if two users are posting similar UGCs, e.g., text content, posting time or location, on two of their OSNs accounts, then these two accounts are likely to belong to a single physical person. This data contains abundant information about the daily schedule of the user, like his preferred time of posting and location of posting.
Unlike the profile attributes like name, description, etc. which can be changed at any time at the users will, the UGCs contain some immutable attributes which cannot be altered by the user once it has been posted, e.g., timestamp. 

We focused on the posting times of the user on different social networks, as the UGC posting time is indicative of a user's work and free time \cite{Li2018a}. For example, a user may be using a professional social network like LinkedIn at their workplace, but, posting on personal OSNs like Facebook or Flickr is unlikely from the workplace. A user has no restrictions to use personal OSNs like Facebook and Flickr from anywhere else but might not be really interested in posting on professional OSNs. 

\begin{figure}[H]
    \centering
    \includegraphics[width=\columnwidth]{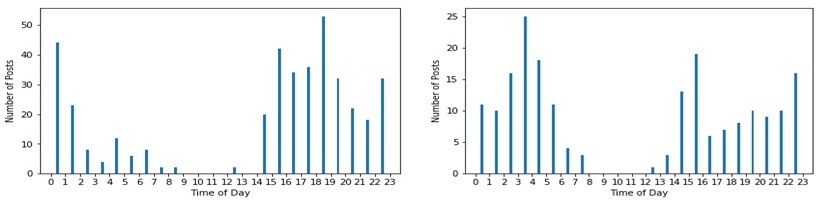}
    \caption{Flickr posts (left) and tweets (right) based on time of the day}
    \label{fig:temporal_flickr_posts_tweets}
\end{figure}

One user posts on multiple OSNs at the same time. While some users share posts/UGCs in the morning, others prefer posting their activities at night. Observing these patterns can be quite helpful for user identification on multiple OSNs. For a user, we can extract the posting time from his public UGCs and obtain a sequence of posting time. We represented the cumulative posting time of UGCs by a user in a vector of hours in a day and days in a week. 
For instance, Figure~\ref{fig:temporal_flickr_posts_tweets} displays the amount of UGCs posted on Flickr and Twitter by the same user, each bar represents one hour of the day. This posting pattern is a unique signature for this user and shows a similar pattern across both OSNs.

Jaccard Similarity can be used to find the similarity between any two sets of numbers, words or characters. In the given case, we calculated the Jaccard similarity of the vectors created above. Jaccard similarity needs similar elements in the vectors but since the vectors consisted of integers computing Jaccard similarity was not very feasible. To overcome this issue, we converted the vectors in a Boolean form were if the user has posted anything in a certain hour or day then the element is set as True, whereas if there is no activity then the element is set to False.

\subsection{Profile Embedding Model - User Identification based on Profile Text Embedding}
\label{sectProfileEmbedModel}

Embeddings are representation of documents/words/character in a vector space, they can be described by an array of numbers after embedding. Each word/character is defined as a vector of some dimension. Unlike the similarity measure used in Section \ref{profile_similarity_model}, word embeddings are capable of capturing context of a word in a document, its semantic and syntactic similarity and relation with other words. 
As word embedding only works for previously seen words, we also utilized character embeddings that are able to generate vectors for out-of-vocabulary words. This capability is important as most usernames we observe are the names of the users, random combination of letters or numbers. We employed word and character level embeddings on the text features present in the users' profiles on both the social networks. This helped us to get a better picture of the textual content

For our study, we used a set of pre-trained embedding models from an open source project called Flair\cite{Flair2018}. Flair is a state of the art embedding tool which allows use to generate different kind of embeddings trained on a variety of sources like news, Wikipedia, Twitter, to name a few. Flair embeddings are trained context based embeddings meaning that they are aware of the relationship between the surrounding words or the words that are dependent on each other.

We exploited the embeddings further by stacking them, i.e., combining the word and character level embeddings to improve the performance of our user identification model. Stacked together, the word and character embeddings are capable of capturing the relation between the words and the patterns of typos and spelling mistakes made by a certain user. 
Embeddings expand the feature set and therefore require a denser neural network to work. In our experimentation, we utilized various configurations of neural networks to find the best fit for the stacked embeddings.

\section{Neural Network Architecture}

The profile similarity, temporal and embedding models proposed in Section~\ref{sectProposedModels} result in three set of features based on various profile similarity measures, UGC posting patterns and textual embedding on user profiles. Using these features as input, we utilized a simple multi-layer perceptron architecture with 1 input layer, 3 hidden layers and 1 output layer for training and testing these models.
Since it is a feed-forward neural network, each layer is fully connected to the subsequent layers. The input layer consists of a varying number of nodes based on the number of inputs, hidden layers comprise of 50 nodes per layer for profile similarity model and temporal model, but it consists of 300 nodes per layers for the profile text embedding model. The figure of 300 nodes was selected by optimising the neural network to work with embeddings, as embeddings can have a lot more features than similarity metrics. Apart from that, the output layer consists of 2 nodes in each case as we are treating the user identification as a classification problem. 

We used the Rectified Linear Unit (ReLU) as our activation function at the input and hidden layers of the neural network.
The ReLU activation function returns 0 if the input value is negative, whereas for any positive value that same value is returned \cite{Glorot2011}. Mathematically, ReLU can be defined as ${f(x)=x^{+}=\max(0,x)}$. ReLU is a computationally cheap, non-linear activation function. One of the benefits of ReLU is that when using random weights for initialising the neural network, almost half of the neurons are not activated as their output would be less than 0. This results is a sparse and light weight neural network. 

At the output layer, we used the softmax activation function to map the non-normalised output of the neural network to a probability distribution over the given classification classes \cite{Duan2007}. Softmax is an activation function that calculates the normalised probability distribution of a vector consisting of K elements such that the sum of the probabilities is one. All the v output values of softmax function ranges from 0 to 1, the probability value of the target class would be more that other classes. Softmax can be used in any of the layers of the neural network and is especially beneficial in the output layer at the time of classification due to its sum of probabilities property.

\begin{figure}[H]
    \centering
$\sigma ({x})_{i}={\frac {e^{x_{i}}}{\sum _{j=1}^{K}e^{x_{j}}}}$    for $i = 1  \dots K$ and $j = 1 \dots K$
    \label{fig:softmax_function}
\end{figure}

where, $x_i$ denotes the $i^th$ element of the vector $X$ and i,j ranges from 1 to $K$, where $K$ is the length of $X$.

We employed Adam optimiser in our neural network, which is based on the moment optimiser~\cite{Kingma2014}. It enhanced the moment optimiser by making it dependent on the previous gradient, so that it can update the learning rate after each iteration. Hence, the name Adam which stands for Adaptive Moment Estimation. Adam is one of the most widely used optimiser as it uses the knowledge from the past gradient to compute the present gradient values by exploiting the concept of moment estimation. This is done by calculating the current gradient by adding a fraction of previous gradients. 
We used Categorical Cross Entropy (CCE) loss as our loss function.

Overfitting is a grave problem in deep neural networks. To tackle overfitting, a dropout can be used. The idea is to randomly drop nodes and their associated connections in the training phase of the deep neural network \cite{Dropout2014}. Dropout inhibits the nodes from adapting too much from the data set. Here, between each of the layers of the model, a dropout of 50\% is implemented.
We also employ early stopping \cite{Prechelt2012} to determine the appropriate number of training iterations to run before the model begins to over-fit.

\section{Experiment Methodology}
\label{data_description}
In order to identify identical users on multiple social networks as a supervised learning problem~\cite{Wang2007}, we require prior knowledge about the users that are available on both social networks as a ground truth label. This section explains the different methods we used to acquire this user data from various OSNs.

\subsection{Social Media Data Collection}

Twitter offers an application programming interface (API) allows us to collect tweets based on a sample stream, keywords or geo-location, subjected to various collection limits \cite{Kerg2014,Roedler2017}. Due to its public availability, Twitter data has been used for many research works in recent years~\cite{de2021and,Kwan-IUI21,Liu-BigData20}. In this work, we used the keywords stream to collect tweets containing the names of various social networks, which are LinkedIn, Instagram, Flickr, Reddit, Tumblr and Pinterest. As user profile information is useful for our work, we further fetched the user information of the corresponding user of each tweet. In particular, we utilize the website address provided on the user profile, which typically correspond to the other OSN accounts of this user. Using this method, we collected 75,398 user pairs from the six OSNs as shown in Table~\ref{fig:OSN_pie_chart}. 

\begin{table}[H]
	\renewcommand*{\arraystretch}{1.1}
\centering
\caption{Number of social network URLs collected}
\label{fig:OSN_pie_chart}
\begin{tabular}{ccc}
\hline
Sno & Social Network & URLs Collected \\
\hline
1 & LinkedIn & 5,866 \\
2 & Instagram & 39,316 \\
3 & Flickr & 1,174 \\
4 & Reddit & 536 \\
5 & Tumblr & 28,411 \\
6 & Pinterest & 631 \\
\hline
\end{tabular}
\end{table}

Based on Table~\ref{fig:OSN_pie_chart}, Instagram-Twitter user pairs had the highest frequency thus we further retrieved the corresponding user profile information, recent posts and comments from Instagram. We performed a manual verification and found a large proportion of Instagram data to be missing due to private profiles, change of usernames, and a transitional phase in the API. As such, we explored other OSNs which could be used for current purposes, namely Flickr due to its good quality and stable API. Similar to Twitter, Flickr data has been frequently used for numerous research works~\cite{Ho-IUI21,Halder-PAKDD21,zhou2020semi}.

Apart from user profile data, we also collected user-generated content (UGC), such as the text content, timestamps, and locations associated with social media postings, as these UGCs allow us to identify the same user across multiple OSNs~\cite{Li2018a}. We further augmented this dataset by using the Flickr-Twitter user pairs provided in \cite{Perito2011,Cosnet2015}. These user pairs serve as ground truth labels for the subsequent training and evaluation. In total, we curated 125,586 tweets and 71,784 Flickr posts from 955 user pairs.

\subsection{Synthetic Dataset Augmentation}

The user information downloaded from the respective social networks contained abundant information which was essential for the current model, hence, a subset of attributes was utilised in the PS model. The users' information collected from Twitter and Flickr was matched based on the user name pairs from \cite{Cosnet2015}, this reduced the dataset to further to 918 valid user pairs.

As our current dataset only comprises positive samples (i.e., user pairs), we require negative samples (i.e., non user pairs) for training a supervised learning model. To achieve this, we performed under-sampling \cite{Goswami2005} by creating random false user pairs and empirically experimented different true-false user pair ratios before selecting a ratio of 1:8 true-false pairs. Using this dataset, we further split 75\% for training and 25\% for evaluation purposes.

\section{Evaluation and Metrics}

We evaluated the algorithms by using 10-fold cross validation strategy, and defined the different possible output based on actual values and predicted values~\cite{terminology_2019} as follows:

\begin{itemize}
    \item \textbf{True Positive (TP):} User-pairs classified as same individual and were same individual in reality.
    \item \textbf{False Positive (FP):} User-pairs classified as different individuals, but in reality, they were the same individual.
    \item \textbf{False Negative (FN):} User-pairs classified as different individuals and were different individuals in reality.
    \item \textbf{True Negative (TN):} User-pairs classified as same individual, but in reality, they were the different individuals.
\end{itemize}

To evaluate the performance of each method, we used the following evaluation metrics:

\begin{itemize}
    \item \textbf{Precision:} Precision is the ratio of correctly predicted user-pairs (TP) to the total user-pairs predicted as same individual (TP+FP) user-pairs, defined as: $Precision = {\frac{TP}{TP+FP}}$.

    \item \textbf{Recall:} Recall is the ratio of correctly predicted user-pairs to all the user-pairs that actually belong to the same individual, defined as: $Recall = {\frac{TP}{TP+FN}}$.
 
    \item \textbf{$\mathbf{F_1}$-Score:} F1 Score is the weighted/harmonic average of Precision and Recall, defined as: $F_1 Score = {2*\frac{Recall * Precision}{Recall + Precision}}$.
    
\end{itemize}

\section{Experiments and Results}
\label{sectExperimentResults}

In this section, we discuss the experimental results and our observations from evaluating our three proposed models.

\subsection{Evaluation of the Profile Similarity Model}
Table \ref{tab:similarity_measures_performance} shows the comparison among the various similarity measures used to identify the user on the two OSNs, i.e., our Profile Similarity model. The table show the performance of the classifier for the True class, i.e. the class where the user-pair is identified as the same on both the networks. Editex similarity outperforms all the similarity measures in comparison with 0.95 F1-score, this is a reasonable outcome as the Editex similarity compares the strings based on the way they sound. This property of Editex plays a major role in this model as the users opt for similar sounding and spelled usernames, therefore, Editex is able to compare the usernames with more accuracy than other similarity measures.

Table \ref{tab:similarity_measures_wo_name_performance} displays the performance of the Profile Similarity Model without training the model on the name based features like the username and the realname. This shows a performance drop of more than 50\% for all the similarity measure signifying the importance of name based features in the process of identification of users online. The NCD based on Bzip2 algorithm performed the best in this case instead of Editex because they used different kind of techniques to compare the text present in the different features. NCD based on Bzip2 algorithm scored 0.40 F1-score for the True class, which is marginally greater than the rest of the algorithms.

\begin{table}[H]
	\renewcommand*{\arraystretch}{1.1}
    \centering
    \caption{Performance of the profile similarity model using various similarity measures performed on all the features}
    \label{tab:similarity_measures_performance}
    \begin{tabular}{cccc}
    \hline
    Similarity & Precision & Recall & F1-Score\\
    \hline
    Levenshtein & 0.95 & 0.90 & 0.93 \\
    Damerau Levenshtein & 0.96 & 0.90 & 0.93 \\
    Editex & \textbf{0.99} & \textbf{0.91} & \textbf{0.95} \\
    Jaro & 0.92 & 0.90 & 0.91 \\
    Jaccard & 0.98 & 0.90 & 0.94 \\
    NCD (Bzip2) & 0.97 & 0.90 & 0.94 \\
    LC Sequence & 0.96 & 0.89 & 0.93 \\
    Smith Waterman & 0.98 & 0.90 & 0.94 \\
    Cosine & \textbf{0.99} & 0.90 & 0.94 \\
    \hline
    \end{tabular}
\end{table}

\begin{table}[H]
	\renewcommand*{\arraystretch}{1.1}
    \centering
    \caption{Performance of the profile similarity model using various similarity measures performed on all the features except the Name based features.}
    \label{tab:similarity_measures_wo_name_performance}
    \begin{tabular}{cccc}
    \hline
    Similarity & Precision & Recall & F1-Score\\
    \hline
    Levenshtein & 0.70 & 0.25 & 0.37 \\
    Damerau Levenshtein & 0.74 & 0.27 & 0.39 \\
    Editex & 0.79 & 0.26 &      0.39 \\
    Jaro & 0.62 &     0.27 & 0.38 \\
    Jaccard & 0.73 &     0.26 & 0.38 \\
    NCD (Bzip2) & 0.66 & \textbf{0.28} & \textbf{0.40}  \\
    LC Sequence & 0.63 & \textbf{0.28} & 0.38 \\
    Smith Waterman & 0.73 & 0.26 & 0.39 \\
    Cosine & \textbf{0.80} & 0.26 & 0.39\\
    \hline
    \end{tabular}
\end{table}

\subsection{Evaluation of Temporal Model}
Table \ref{tab:ugc_based_temporal_model_performance} shows the performance of the Temporal Model, i.e., temporal features derived from UGC posting patterns. The two methods used in this approach are posting times based on Day of Week (DOW) and posting times based on Hour of Day (HOD). Out of these two features (HOD) performed slightly better, this reflects the effect of more granularity in the HOD representation of time. 

\begin{table}[H]
	\renewcommand*{\arraystretch}{1.1}
    \centering
    \caption{Performance of the Temporal Model based on UGC Postings.}
    \label{tab:ugc_based_temporal_model_performance}
    \begin{tabular}{cccc}
    \hline
    Method & Precision & Recall & F1-Score\\
    \hline
    Hour of Day & \textbf{0.87} & \textbf{1.00}  & \textbf{0.93}\\
    Day of Week & 0.85 & 1.00 & 0.92\\
    \hline
    \end{tabular}
\end{table}

\subsection{Profile Embedding Model}

Table \ref{tab:embedding_without_desc_performance} shows the comparison of the performance of the Profile Embedding model without training the neural network on the profile description of the users. Since this feature set comprised of name based features, it was able to perform better with 50 nodes per hidden layer with F1-score of 0.81, 0.91 precision but with a lower recall of 0.69. In contrast, the variation with 400 nodes had the best recall of 0.82. 

\begin{table}[H]
	\renewcommand*{\arraystretch}{1.1}
    \centering
    \caption{Performance of the Profile Text Embedding model Excluding Description.}
    \label{tab:embedding_without_desc_performance}
    \begin{tabular}{cccc}
    \hline
    Nodes & Precision & Recall & F1-Score\\
    \hline
    50 &  \textbf{0.98} & 0.69 & \textbf{0.81}\\
    100 & 0.96 & 0.69 & 0.80\\
    200 & 0.36 & 0.75 & 0.48\\
    300 & 0.37 & 0.80 & 0.51\\
    400 & 0.38 & \textbf{0.82} & 0.52\\
    \hline
    \end{tabular}
\end{table}

The performance of the model with the description embeddings included in the training was quite similar to the one without the description embeddings with the best F1-Score of 0.81 for 200 and 300 nodes per layer, as shown in Table \ref{tab:embedding_with_desc_performance}. We also observed precision scores to be slightly better than the previous model at 0.99 for the 50 nodes variation. However, the maximum recall we got from both models with and without description training came out to be 0.82 for the 400 nodes per layer variation.

\begin{table}[H]
	\renewcommand*{\arraystretch}{1.1}
    \centering
    \caption{Performance of the Profile  Embedding model Including Description.}
    \label{tab:embedding_with_desc_performance}
    \begin{tabular}{cccc}
    \hline
    Nodes & Precision & Recall & F1-Score\\
    \hline
    50 &  \textbf{0.99} & 0.64 & 0.77\\
    100 &  0.92 & 0.71 & 0.80\\
    200 &  0.97 & 0.70 & \textbf{0.81}\\
    300 &  \textbf{0.99} & 0.69 & \textbf{0.81}\\
    400 &  0.37 & \textbf{0.82} & 0.51\\
    \hline
    \end{tabular}
\end{table}

\subsection{Main Findings and Observations}

Based on the experimental results from the Profile Similarity Model, Temporal Model and Profile Embedding Model, we make the following main observations:

\begin{itemize}
\item In the Profile Similarity model, incorporating user names and real names results in a three-fold improvement in Recall, two-fold improvement in F1-score and a smaller but noticeable improvement in Precision. This result shows the importance of considering user names and real name in any user identification model, and how this improvement is independent of the similarity measure used.
\item In the Temporal Model, we aim to identify similar users based on their posting times of UGC and found that representing posting times by HOD works better than DOW. This result show that user posting patterns are better represented by the daily patterns (hour of day), rather than the weekly posting patterns (day of week).
\item In the Profile Embedding model, we observe that the model that considers profile description performs similarly to the same model that does not consider profile description, except Precision where the former performs slightly better. This result shows that profile description does not play a major role in identifying similar users across OSNs.
\end{itemize}

\section{Conclusion and Future Work}

In this paper, we study the problem of user identification across multiple OSNs. To address this problem, we propose three models, namely: (i) Profile Similarity model based on various user profile features coupled with similarity measures; (ii) Temporal Model based on temporal patterns in the posting of UGCs; and (iii) Profile Embedding model that utilizes word/character embedding techniques on textual features in the user profile. All models are then fed into a multi-layer perceptron to classify whether a user pair correspond to the same user or not. Our results show that the Profile Similarity model performs the best, which indicate that simply using user profile features based on similarity measures work well with the user name and real names playing a key role in user identification. The Profile Embedding model comes in a close second, which highlights the ability of this model in capturing the meaning and the relation between the words and characters of the text present in the profiles. While the Temporal model performs the worst among the three models, the former still offers good performance in terms of Precision, Recall and F1-score, which indicates the usefulness of individual posting patterns in user identification.

There are also various possible directions for future work. Firstly, we can combine the three models to best utilize the usefulness of profile similarity, temporal patterns and profile description embeddings. Secondly, we can explore other models for addressing this task, such as by using additional similarity measures, e.g., overlap index, and utilizing other neural network architectures, e.g., siamese networks~\cite{mueller2016siamese}. Lastly, we can incorporate more demographics related information into our models, such as considering the cultural aspects of a user, which may affect his/her willingness to disclose certain information.

\vspace{2mm}
{\small
{\noindent\bf Acknowledgements}. We thank the reviewers for their useful comments. This research is funded in part by the Singapore University of Technology and Design under grant SRG-ISTD-2018-140.
}

\balance
\bibliographystyle{IEEEtran}
\bibliography{multiOSN-IJCNN21}

\end{document}